\documentclass{article}
\usepackage{spconf,amsmath,graphicx}
\usepackage{booktabs}

\usepackage{graphicx}
\usepackage{svg}
\usepackage{xcolor}
\usepackage{hyperref}
\usepackage{amssymb}
\usepackage{siunitx}
\usepackage{caption}
\usepackage{subcaption}
\usepackage{float}

\usepackage{enumitem}
\setlist{nosep, leftmargin=14pt}

\usepackage{mwe} 

\title{Attention-guided classification and tumor localization of combined hepatocellular-
cholangiocarcinoma}

\name{%
\begin{tabular}{c}
Xiaowen Liang$^{1}$, Hélène Cazier$^{2}$, Aurélie Beaufrère$^{2,3}$\\
Miguel Albuquerque$^{2,3}$, Valérie Paradis$^{2,3}$, Cédric Wemmert$^{1}$
\end{tabular}
}

\address{%
\begin{tabular}{c}
$^{1}$Université de Strasbourg, CNRS, ICube UMR 7357, Illkirch, France\\[2pt]
$^{2}$Université Paris Cité, INSERM, UMR-S1149, Centre de Recherche sur l’Inflammation (CRI),\\
Laboratoire d’Excellence Inflamex, F-75018 Paris, France\\[2pt]
$^{3}$Pathology Department, Hôpital Beaujon, AP-HP, Clichy, France
\end{tabular}
}

\begin{document}
%
\maketitle
\begin{abstract}
Combined hepatocellular-cholangiocarcinomas (cHCC-CCA) are a type of cancer that share components of two other primary liver cancers: hepatocellular carcinoma (HCC) and cholangiocarcinoma (CCA). Accurate diagnosis of cHCC-CCA, of which the behavior may resemble that of either HCC or CCA, remains a challenge due to strong intratumoral heterogeneity. Since the prognoses differ between liver cancer types, it is necessary to improve their morphological characterization and recognition in order to accurately identify cHCC-CCA. In this work, we use whole slide images (WSIs) of the two main types of liver cancer (HCC and CCA) and cHCC-CCA, stained with Hematoxylin–Eosin–Safran (HES), to train a classification model that can distinguish between these subtypes and provides a potential solution for localizing different types of tumors in the combined liver cancer images.
\end{abstract}
\section{Introduction}
\label{sec:intro}

Hepatocellular carcinoma (HCC) and cholangiocarcinoma (CCA) are the major primary liver carcinomas, whereas combined hepatocellular-cholangiocarcinoma (cHCC-CCA) exhibits both hepatocytic and cholangiocytic differentiation~\cite{brunt2018chcc}. This morphological heterogeneity motivates computational approaches for both whole-slide image (WSI) classification and component localization. Calderaro \textit{et al.}~\cite{calderaro2023deep}  used a classifier trained on pure HCC and intrahepatic CCA (iCCA) to characterize HCC-like and iCCA-like morphology in cHCC-CCA. Beaufr{\`e}re \textit{et al.}~\cite{beaufrere2024primary} combined supervised tumor detection with feature clustering to estimate HCC and iCCA component proportions in cHCC-CCA biopsies.

Since detailed WSI annotations are difficult to obtain, weakly supervised multiple instance learning (MIL) performs classification from slide-level labels only. ABMIL~\cite{ilse2018attention} introduced attention-based instance aggregation, which CLAM~\cite{lu2021data} and TransMIL~\cite{shao2021transmil} extended through instance-level clustering and inter-instance modeling, respectively. MS-CLAM~\cite{tourniaire2023ms} further incorporated limited patch-level annotations and attention supervision to improve tumor classification and localization. However, high classification accuracy does not guarantee pathologically plausible attention or faithful explanations of model predictions~\cite{idaji2026beyond}.

We propose an attention-supervised three-class MIL model for HCC, CCA, and cHCC-CCA classification, using attention references from a binary HCC–CCA model to encourage pathological plausibility. We also propose a jointly trained dual-branch model that learns complementary attention patterns within tumor regions, generating distinct heatmaps of the same cHCC-CCA slide to localize HCC-like and CCA-like components.

\section{Methodology}
\label{sec:methodology}

\subsection{Attention-guided three-class classification}
\label{sec:3-class}

\begin{figure}[htb]
\centering

{\includegraphics[width=8.5cm,trim={0cm 3cm 0cm 0cm},clip]{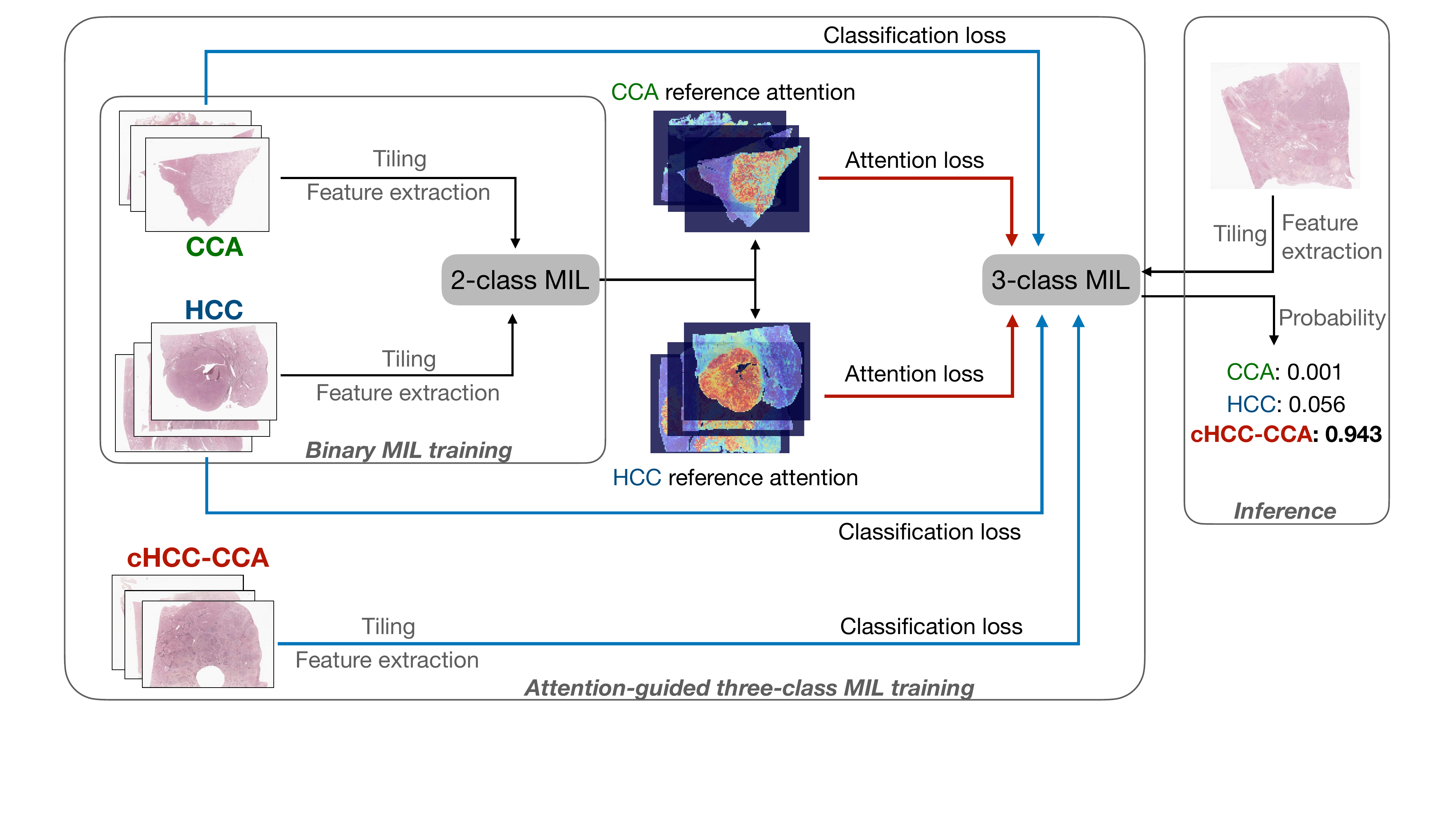}}

\caption{Attention-guided three-class MIL training.}
\label{fig:attention_supervision}
\end{figure}

Attention-based MIL assigns weights to instances within a bag, enabling the visualization of potentially relevant tissue regions.
In our experiments, introducing cHCC-CCA as a third class alongside HCC and CCA sometimes shifted attention toward
regions inconsistent with pathological annotations, indicating a low pathological plausibility of the attention maps. We therefore use attention maps generated by a binary HCC-CCA model as fixed references to supervise the three-class model on HCC and CCA slides (Figure~\ref{fig:attention_supervision}). We adopt TransMIL~\cite{shao2021transmil}, which contains
two self-attention blocks separated by a Pyramid Position Encoding Generator (PPEG). Patch features are projected
into $D$-dimensional embeddings and augmented with a learnable class token (CLS). Each attention block uses
$H$ heads of dimension $D/H$. We extract the CLS-query attention vectors from both blocks:

\[
  A^{(\ell)} \in \mathbb{R}^{H \times N},
  \qquad \ell \in \{1,2\},
\]
where $N$ is the number of tokens, including the class token. To obtain a class-specific attention map, the classifier weights are used to estimate the importance of each attention head.
For class $c$, the score of head $h$ is computed as
\begin{equation}
  s_{c,h} = \sum_{d=1}^{D/H} z_{h,d} W_{c,h,d},
\end{equation}
where $z_{h,d}$ denotes the $d$-th component of the representation
of head $h$, and $W_{c,h,d}$ denotes the corresponding classifier
weight for class $c$. Different heads are fused in an attention-weighted way in our study, the head scores are normalized with softmax and used as fusion weights:
\begin{equation}
  \alpha_{c,h} = \frac{\exp(s_{c,h})}{\sum_{j=1}^{H}\exp(s_{c,j})},
  \qquad
  \bar{A}^{(\ell)} = \sum_{h=1}^{H} \alpha_{c,h} A^{(\ell)}_h.
\end{equation}

For attention supervision, we use the fused student attention $\bar{A}^{(\ell)}_{\mathrm{student},i}$ and a fixed, precomputed teacher reference $\bar{A}^{(\ell)}_{\mathrm{teacher},i}$ normalized to unit sum.
The teacher and student vectors correspond to the same tokens in the
same order. Supervision is applied separately to the two attention blocks.
Their agreement is encouraged through a combination of distribution,
magnitude, and directional matching by KL divergence $\mathcal{L}_{\mathrm{KL}}^{(\ell)}$, L1 distance $\mathcal{L}_{\mathrm{L1}}^{(\ell)}$ and cosine distance $\mathcal{L}_{\mathrm{cos}}^{(\ell)}$.
A relative-attention loss enforces a minimum student-attention
ratio between tokens above the teacher's $q_{\mathrm H}$-quantile
and those below its $q_{\mathrm L}$-quantile:
\begin{equation}
\label{eq:rank_loss}
\mathcal{L}_{\mathrm{rank}}^{(\ell)}
=
\left[
m-
\frac{\mu_{\mathcal H}^{(\ell)}}
{\max(\mu_{\mathcal B}^{(\ell)},\epsilon_r)}
\right]_+,
\end{equation}
where $\mu_{\mathcal H}^{(\ell)}$ and
$\mu_{\mathcal B}^{(\ell)}$ denote the mean student attention
over these two sets, respectively, and $[x]_+=\max(0,x)$.
This term is zero if either set is empty.

The overall objective is
\begin{equation}
\begin{aligned}
\mathcal{L}_{\mathrm{attn}}
  &= \sum_{\ell=1}^{2}\delta^{(\ell)}
     \sum_{k\in\mathcal K}
     \lambda_k\mathcal{L}_k^{(\ell)},\\
\mathcal{L}
  &= \mathcal{L}_{\mathrm{CE}}
     +\lambda_{\mathrm{attn}}\mathcal{L}_{\mathrm{attn}},
\end{aligned}
\end{equation}
where $\mathcal K=\{\mathrm{KL},\mathrm{L1},
\mathrm{cos},\mathrm{rank}\}$ and $\delta^{(\ell)}$
indicate whether reference supervision is available and
enabled for layer $\ell$. CCA and HCC slides receive both classification and attention supervision. cHCC-CCA slides receive classification supervision only. 

\subsection{Dual-branch model for tumor component localization}

\begin{figure*}[t]
\centering

{\includegraphics[width=16cm,trim={0cm 4cm 0cm 1cm},clip]{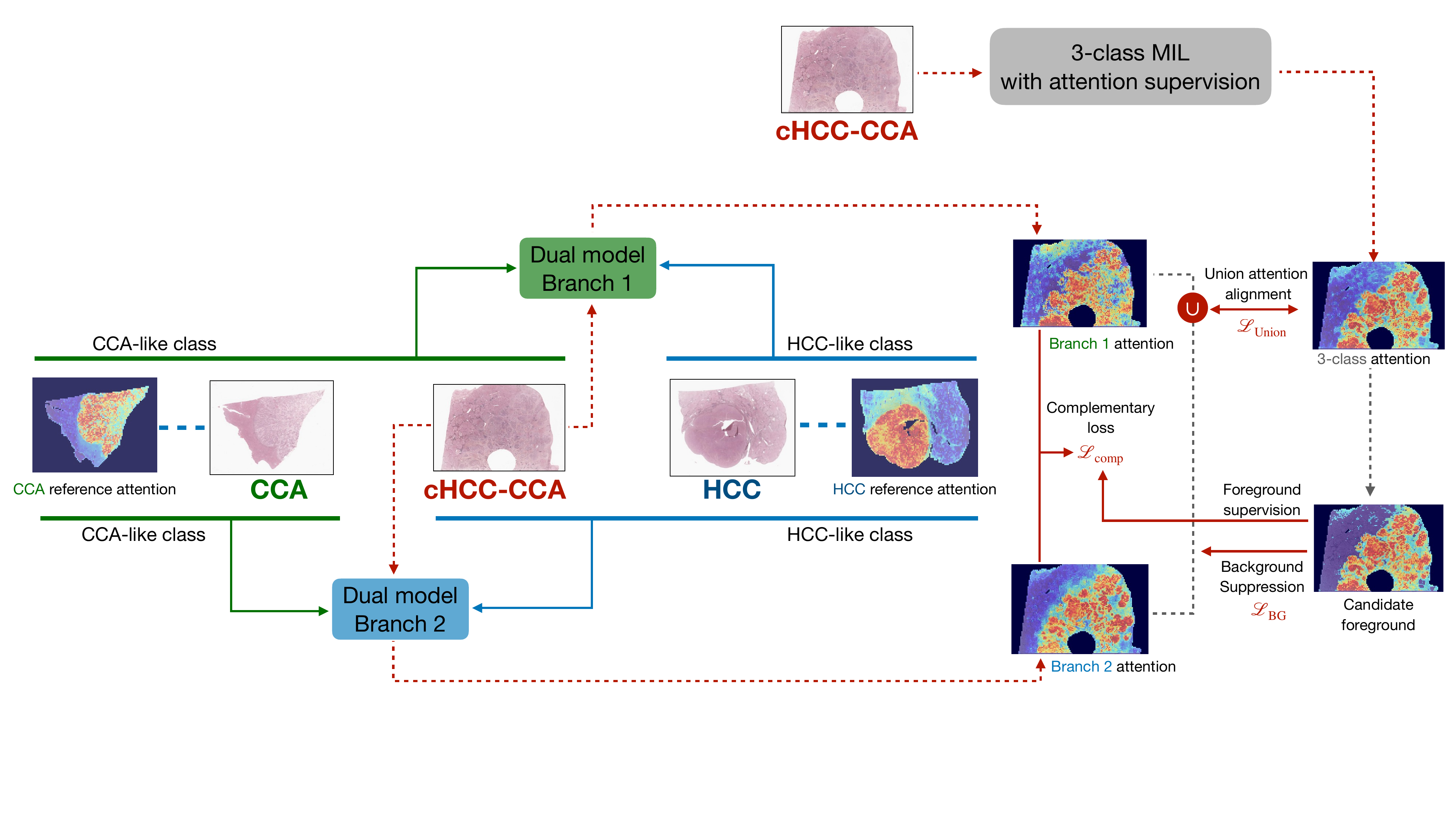}}

\caption{Joint learning of the three-class model and the dual model. The 3-class MIL block is the same as described in Figure~\ref{fig:attention_supervision}. Red dotted lines indicate the generation of cHCC-CCA attention heatmaps used to compute the auxiliary loss $\mathcal{L}_{\mathrm{aux}}$, comprising union alignment loss $\mathcal{L}_{\mathrm{Union}}$, background suppression loss $\mathcal{L}_{\mathrm{BG}}$, and foreground-guided complementary loss $\mathcal{L}_{\mathrm{comp}}$. The three-class model and both branches of the dual model are trained with classification and attention supervision as described previously.}
\label{fig:joint_model}
\end{figure*}

Beyond attention supervision, we encourage complementary attention to HCC-like and CCA-like components in cHCC-CCA tumors. We introduce a dual-branch MIL model jointly trained with the three-class model to generate complementary attention maps for each cHCC-CCA slide. Figure~\ref{fig:joint_model} provides an overview of the joint training framework. The three-class model retains the classification and attention supervision described in Section~\ref{sec:3-class}. The dual model consists of two binary MIL branches. Branches 1 and 2 distinguish CCA/cHCC-CCA from HCC and HCC/cHCC-CCA from CCA, respectively. Both receive classification supervision for all slides and reference-based attention supervision for HCC and CCA slides only. These complementary label groupings are intended to promote CCA-like and HCC-like attention in branches 1 and 2, respectively. To prevent complementarity from shifting attention toward non-tumor regions, we restrict the complementary loss to a candidate foreground. This mask is obtained by combining, normalizing, and thresholding the three-class attention maps from both transformer blocks and is detached from the computation graph.
Background suppression penalizes excessive attention outside the mask, while union alignment encourages the probabilistic union of the dual-branch maps to match the three-class attention distribution.

For cHCC-CCA slides, let $\bar{A}_b^{(\ell)}$ denote the head-fused attention map of branch $b\in\{0,1\}$ at block $\ell$. The foreground mask $M$, with token set $\mathcal{F}$, is obtained using normalization $\mathcal{N}_{\mathrm{vis}}$ and threshold $t_{\mathrm{FG}}$ as described above.
We compute the Pearson correlation $\rho^{(\ell)}$ between the normalized branch maps within $\mathcal{F}$. The complementary loss encourages negative correlation of a target:
\begin{equation}
\mathcal{L}_{\mathrm{comp}}
=
\frac{1}{L}\sum_{\ell=1}^{L}
\left[\rho^{(\ell)}-\rho_{\mathrm{target}}\right]_+, \qquad L=2
\end{equation}
The union and the corresponding three-class cHCC-CCA attention map are normalized into distributions, and their discrepancy $D^{(\ell)}$ is measured using KL divergence.
The background suppression loss penalizes attention exceeding
$\tau_{\mathrm{BG}}$ outside the candidate foreground,
averaged across both branches and all blocks:
\begin{equation}
\mathcal{L}_{\mathrm{BG}}
=
\frac{1}{2L}
\sum_{\ell=1}^{L}\sum_{b=1}^{2}
\frac{
\sum_i (1-M_i)
\left[a_{b,i}^{(\ell)}-\tau_{\mathrm{BG}}\right]_+^2
}{
\max\!\left(1,\sum_i(1-M_i)\right)
}.
\end{equation}


For cHCC-CCA slides, the dual-branch attention loss combines union alignment, foreground complementarity, and background suppression. The loss and overall joint training objective are defined as:
\begin{align}
\mathcal{L}_{\mathrm{dual\text{-}attn}}
&=
\lambda_{\mathrm{union}}\mathcal{L}_{\mathrm{union}}
+
\lambda_{\mathrm{comp}}\mathcal{L}_{\mathrm{comp}}
+
\lambda_{\mathrm{BG}}\mathcal{L}_{\mathrm{BG}},
\\
\mathcal{L}_{\mathrm{joint}}
&=
\mathcal{L}_{\mathrm{3class}}
+
\mathcal{L}_{\mathrm{dual}}
+
\mathcal{L}_{\mathrm{dual\text{-}attn}}.
\end{align}



\section{Experiments}
\label{sec:experiments}

\subsection{Dataset and experimental setup}
We used Hematoxylin--Eosin--Safran (HES)-stained WSIs of HCC, CCA, and cHCC-CCA for model training and validation. The dataset comprised 338 HCC slides, 353 CCA slides, and 222 cHCC-CCA slides. Slides were tiled at $10\times$ magnification, and patch-level features were extracted using the UNI2-h foundation model~\cite{chen2024uni}, yielding 1536-dimensional embeddings as inputs to the MIL models. The binary CCA--HCC MIL model used to generate reference attention maps was trained for 100 epochs. The three-class MIL model and the jointly trained three-class and dual models were trained for a maximum of 50 epochs, with early stopping triggered when the validation classification loss did not improve for 20 consecutive epochs. For the complementary loss, gradient scaling factors were set to 1.0 and 0.5 for the CCA-like and HCC-like branches, respectively. All models were trained and evaluated using patient-level 10-fold cross-validation, ensuring that slides from the same patient were assigned to the same fold. Only slide-level class labels were used for training and validation. The independent evaluation set comprised 183 WSIs: 72 HCC, 62 CCA, and 49 cHCC-CCA slides. It includes 66 slides from the external TCGA dataset, consisting of 36 HCC and 30 CCA slides.

\subsection{Attention-guided three-class model}

Table~\ref{tab:model_comparison} shows the classification performance of the three-class model and the 3-class part of the joint training. The results are the average of 10 folds, generated by the epochs that have the best total loss on the three-class model ($\mathcal{L}_{\mathrm{CE}}+\lambda_{\mathrm{attn}}\mathcal{L}_{\mathrm{attn}}$). The recalls of the dual model are given by the intersection of the predictions of its two branches.

The 3-class model without attention supervision achieved the highest macro F1-score (0.880). Attention supervision slightly reduced this score to 0.872 while improving heatmap plausibility (Table~\ref{tab:tac}), suggesting a trade-off between classification and localization. Joint training partially recovered the macro F1 to 0.876. Although the dual-branch model achieved a lower score (0.834), its main contribution lies in component localization through complementary attention maps.
\begin{table}[h]
    \centering
    \caption{Classification performance of the three-class models (rows 1--3) and the dual-branch model (row 4).}
    \label{tab:model_comparison}
    \begin{tabular}{lcccc}
        \toprule
        Model & CCA & HCC & TM & Macro F1 \\
      & Recall & Recall & Recall & \\

        \midrule
        w/o attn. sup. & 0.915 & 0.868 & 0.867 & \textbf{0.880} \\
        w/ attn. sup. & 0.884 & \textbf{0.871} & \textbf{0.871} & 0.872 \\
        Joint & 0.903 & 0.863 & \textbf{0.871} & 0.876 \\
        Dual & \textbf{0.918} & 0.768 & 0.816 & 0.834 \\
        \bottomrule
    \end{tabular}
\end{table}

\begin{table}[h]
    \centering
    \caption{Attention plausibility on 30 slides per class, without (w/o AS) and with attention supervision (w/ AS). $\overline{\mathrm{TAC}}$ denotes per-slide TAC averaged across the ten fold models.}
    \label{tab:tac}
    \setlength{\tabcolsep}{4pt}
    \begin{tabular}{lcccc}
        \toprule
        Class & \multicolumn{2}{c}{TAC}
              & \multicolumn{2}{c}{Slides with $\overline{\mathrm{TAC}}<1$} \\
        \cmidrule(lr){2-3}
        \cmidrule(lr){4-5}
        & w/o AS & w/ AS & w/o AS & w/ AS \\
        \midrule
        CCA & $3.04 \pm 1.63$ & $5.66 \pm 1.12$
            & $0$ & $0$ \\
        HCC & $1.75 \pm 0.51$ & $6.50 \pm 1.72$
            & $6$ & $0$ \\
        \bottomrule
    \end{tabular}
\end{table}

\begin{figure}[htbp]
    \centering
    \begin{subfigure}[t]{0.32\linewidth}
        \centering
        \includegraphics[width=\linewidth]
        {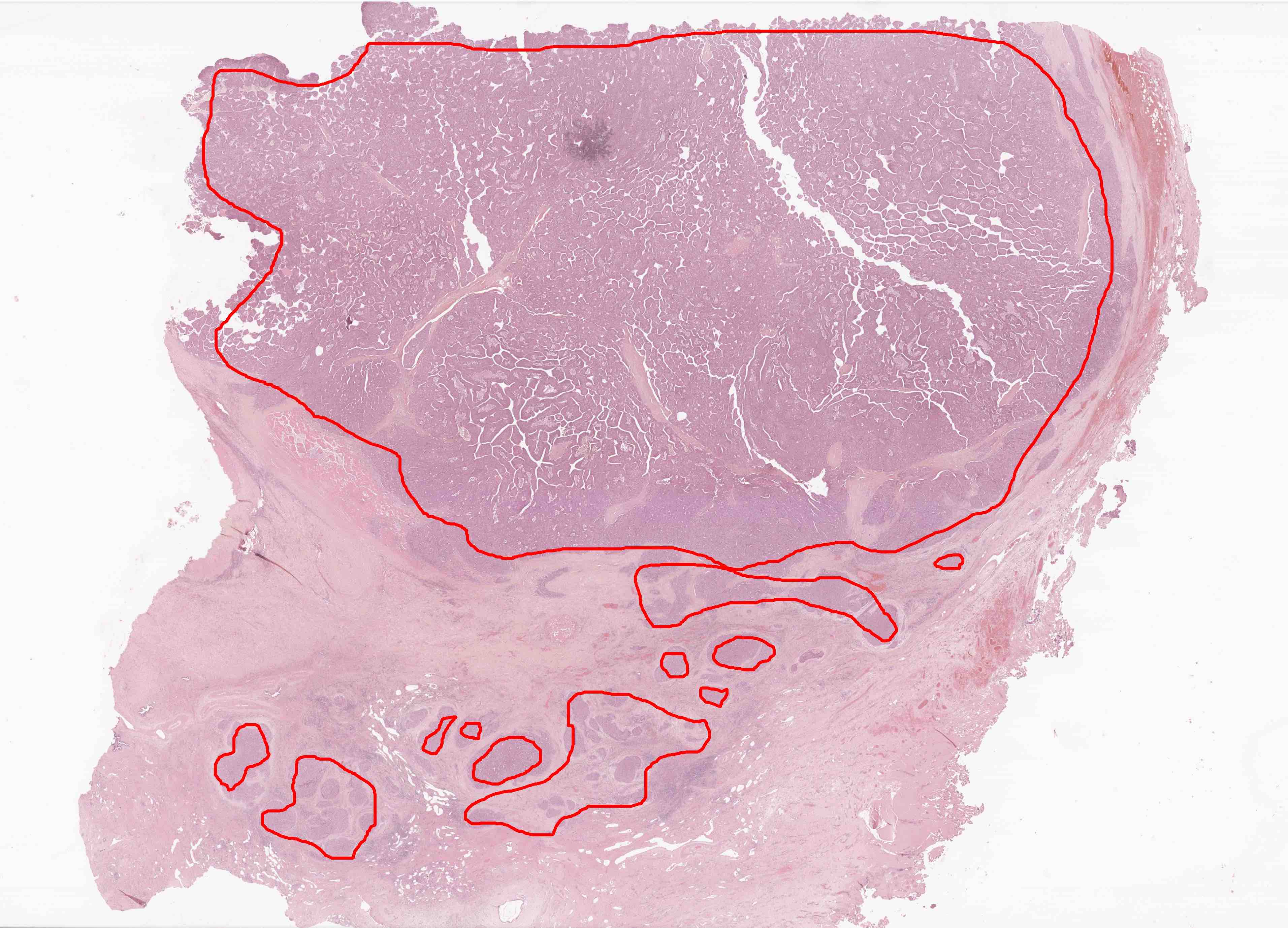}
        \caption{Tumor annotation}
        \label{fig:hcc_annotation}
    \end{subfigure}
    \hfill
    \begin{subfigure}[t]{0.32\linewidth}
        \centering
        \includegraphics[width=\linewidth]
        {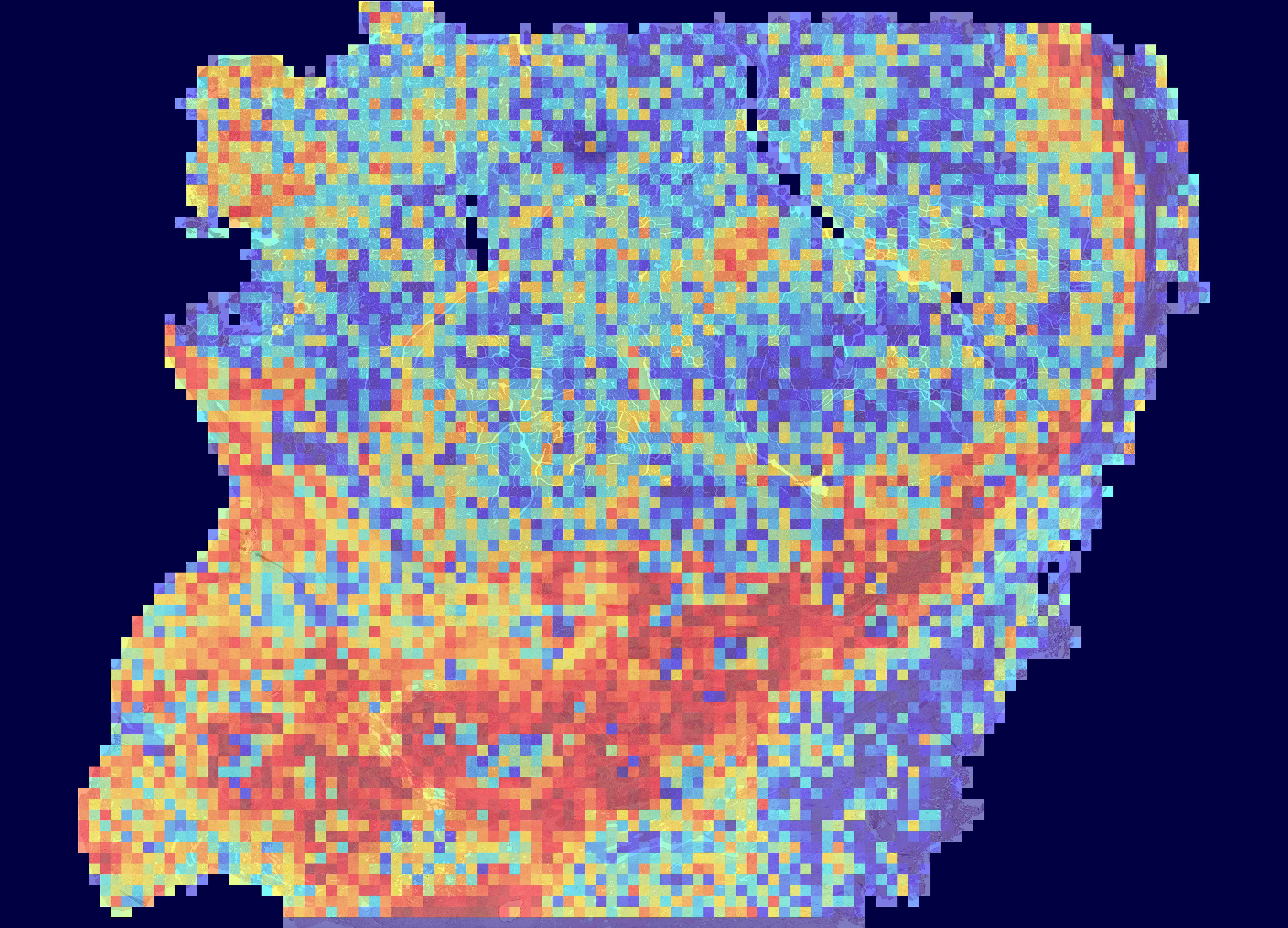}
        \caption{w/o supervision}
        \label{fig:hcc_without_attn}
    \end{subfigure}
    \hfill
    \begin{subfigure}[t]{0.32\linewidth}
        \centering
        \includegraphics[width=\linewidth]
        {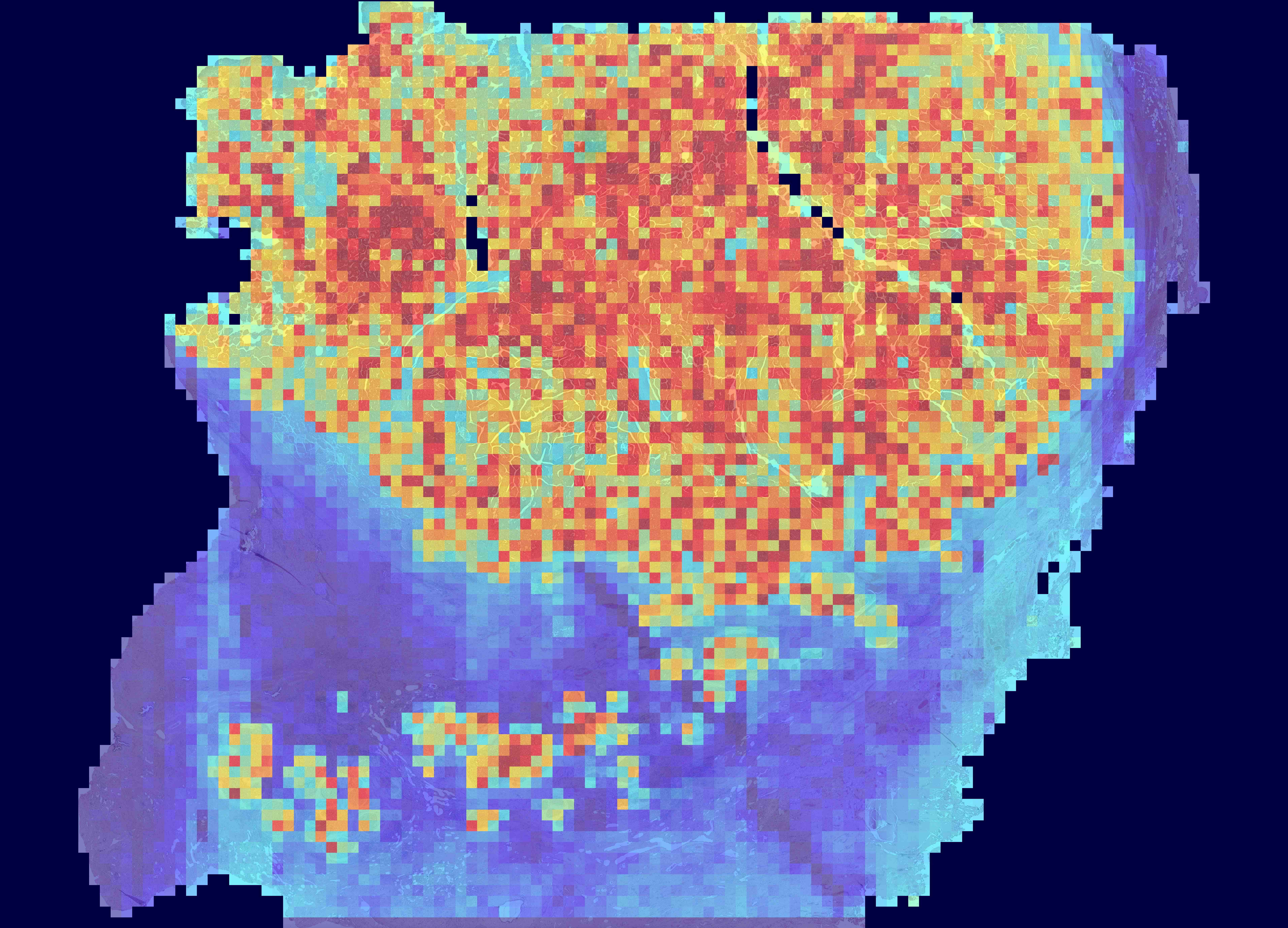}
        \caption{w/ supervision}
        \label{fig:hcc_with_attn}
    \end{subfigure}
    \caption{HCC slide with annotated tumor regions and
    attention heatmaps without and with
    attention supervision.}
    \label{fig:hcc_attention_comparison}
\end{figure}

Heatmap plausibility is assessed using tumor attention contrast (TAC), the ratio of mean normalized attention inside to outside annotated tumor regions. TAC $>1$ indicates preferential tumor attention. Attention supervision increased TAC on 30 CCA and 30 HCC slides (Table~\ref{tab:tac}). Six HCC slides had $\overline{\mathrm{TAC}}<1$ without supervision, all of
which exceeded 1 with supervision. Figure~\ref{fig:hcc_attention_comparison} illustrates one such case.

\subsection{HCC-like and CCA-like component localization}

The dual-branch model is designed to localize and distinguish HCC-like and CCA-like components in cHCC-CCA tumors. We evaluate component selectivity using the CCA/HCC attention ratio for each dual branch and the three-class model:
\begin{equation}
R_b=\frac{\mu_{b,\mathrm{CCA}}}{\mu_{b,\mathrm{HCC}}},
\qquad b\in\{1,2,\text{3-class}\},
\end{equation}
where $\mu_{b,c}$ denotes the mean normalized attention of the model or branch $b$ within regions annotated as component $c$.
Ratios greater than 1 indicate preferential attention to CCA regions, whereas ratios below 1 indicate preferential attention to HCC regions. Branch 1 is intended to highlight CCA-like components, and branch 2 HCC-like components; the three-class model serves as a reference. For dual model heatmap evaluation, we chose the models with the best total loss from the joint training.

\begin{table}[t]
    \centering
    \caption{Component selectivity on 15 annotated cHCC-CCA slides, measured by the CCA/HCC attention ratio and averaged across ten fold models.}
    \label{tab:component_selectivity}
    \begin{tabular}{lc}
        \toprule
        Model & CCA/HCC attention ratio \\
        \midrule
        Three-class model & $0.677 \pm 0.090$ \\
        CCA-like branch & $2.483 \pm 0.873$ \\
        HCC-like branch & $0.480 \pm 0.102$ \\
        \bottomrule
    \end{tabular}
\end{table}

\begin{figure}[htbp]
    \centering
    \begin{subfigure}[t]{0.22\linewidth}
        \centering
        \includegraphics[width=\linewidth]
        {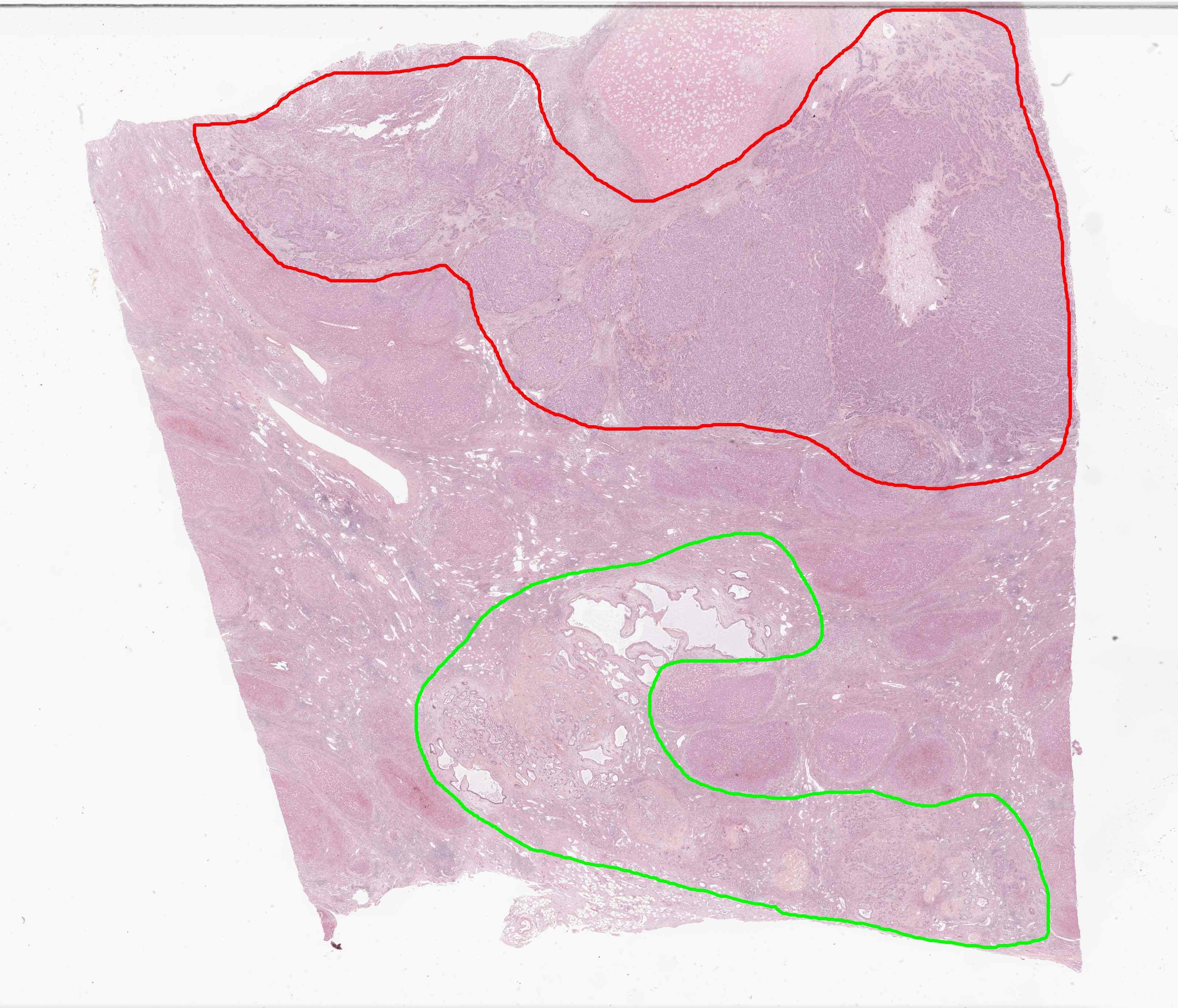}
        \caption{Annot.}
        \label{fig:tm_annotation}
    \end{subfigure}
    \hfill
    \begin{subfigure}[t]{0.22\linewidth}
        \centering
        \includegraphics[width=\linewidth]
        {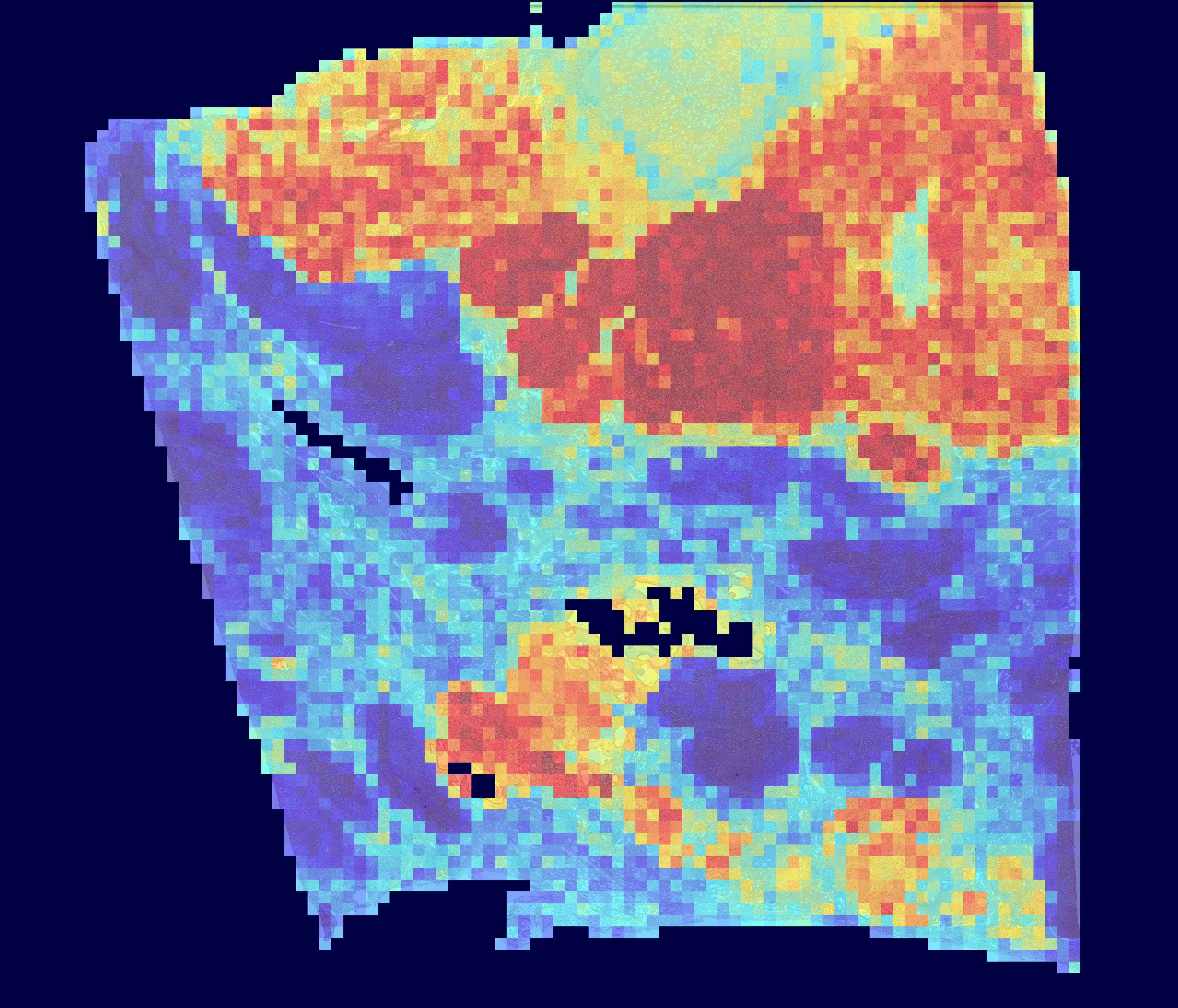}
        \caption{3-class}
        \label{fig:tm_3_class}
    \end{subfigure}
    \hfill
    \begin{subfigure}[t]{0.22\linewidth}
        \centering
        \includegraphics[width=\linewidth]
        {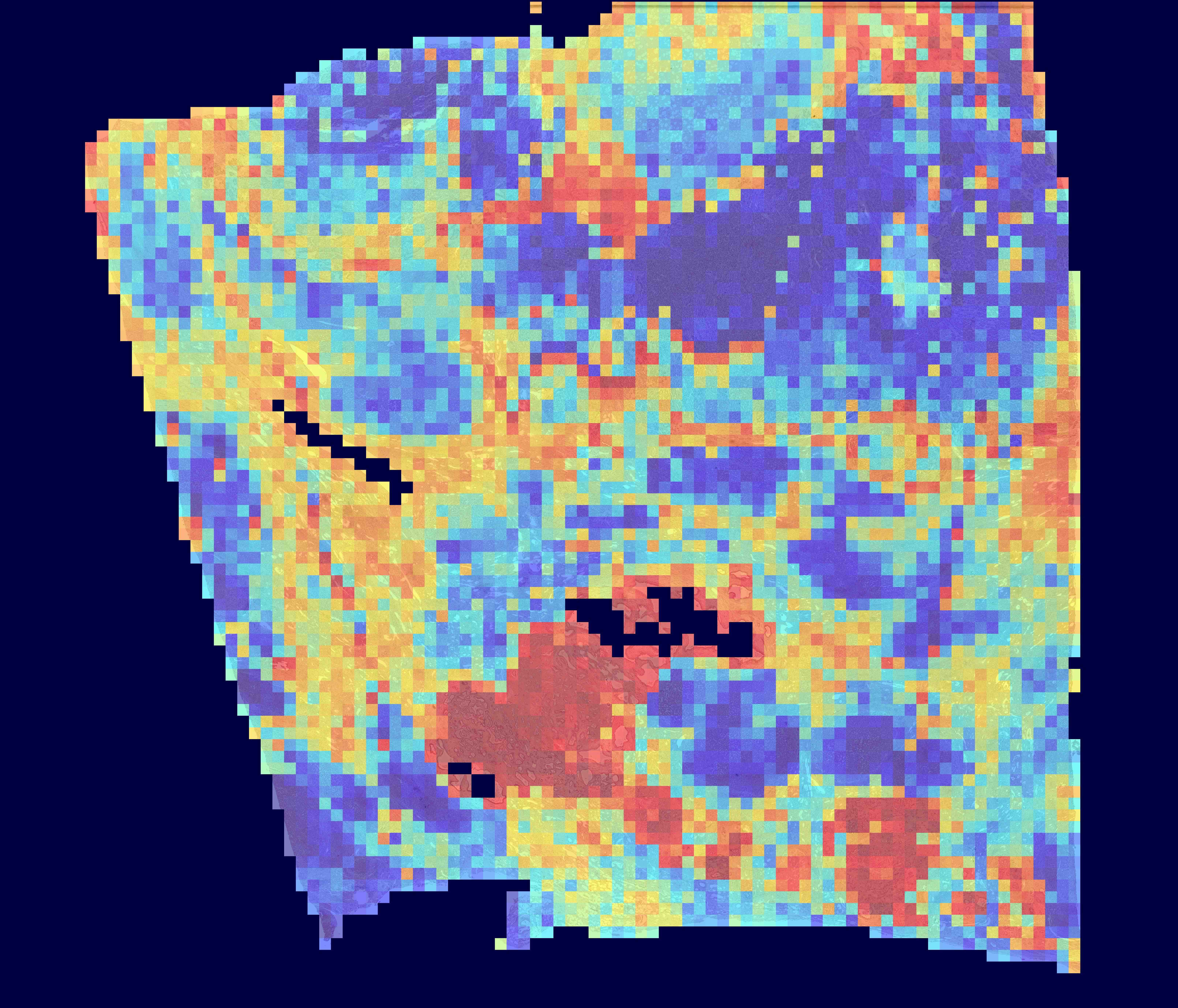}
        \caption{CCA-like}
        \label{fig:tm_branch_1}
    \end{subfigure}
      \hfill
    \begin{subfigure}[t]{0.22\linewidth}
        \centering
        \includegraphics[width=\linewidth]
        {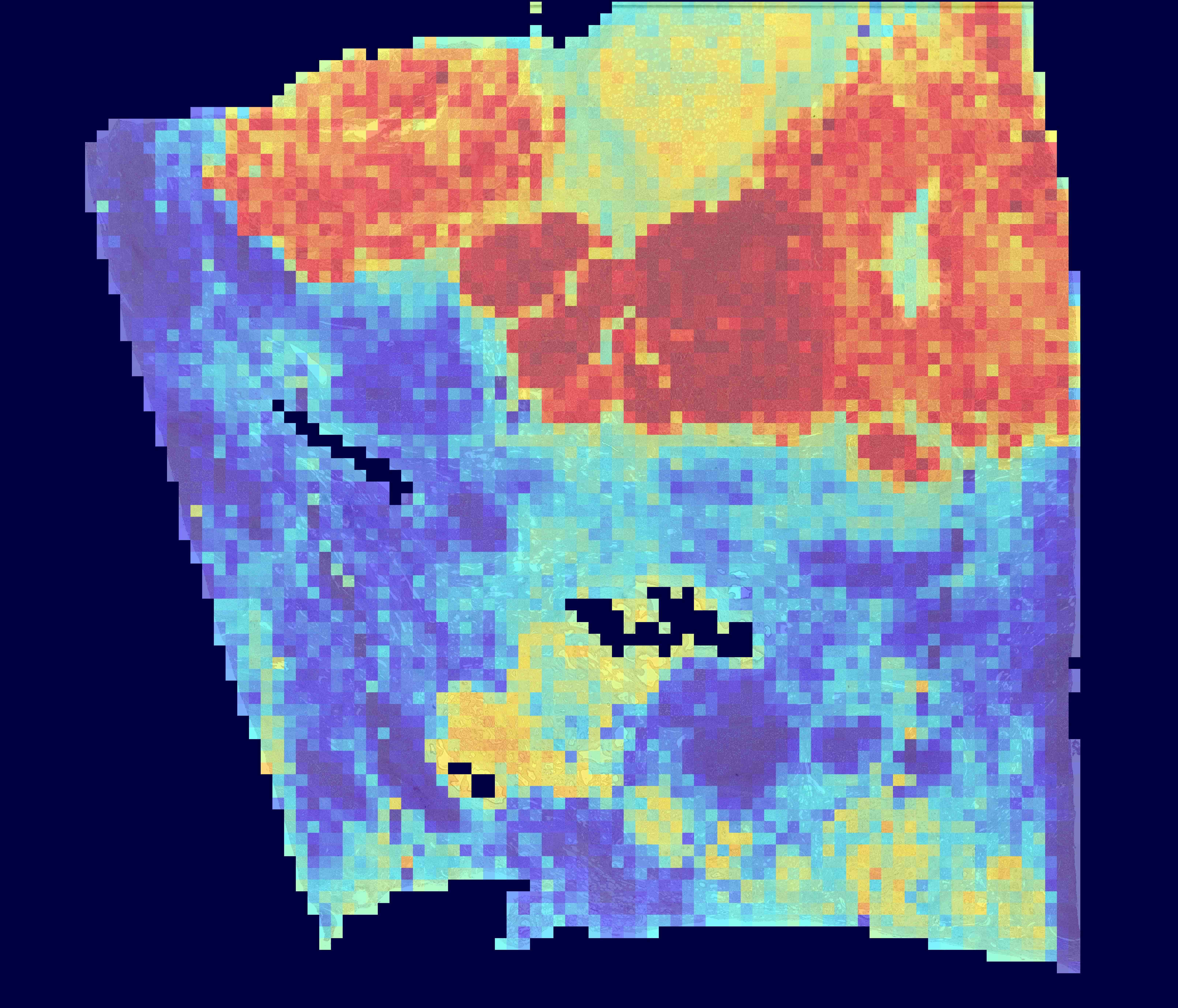}
        \caption{HCC-like}
        \label{fig:tm_branch_1}
    \end{subfigure}
    \caption{cHCC-CCA slide with (a) annotated tumor regions (CCA in green, HCC in red) and attention heatmaps of (b) 3-class model, (c) branch 1 and (d) branch 2.}
    \label{fig:tm_attention_comparison}
\end{figure}

As shown in Table~\ref{tab:component_selectivity}, the CCA-like branch assigns higher mean attention to annotated CCA regions, whereas the HCC-like branch shows the opposite preference. The three-class model also favors HCC regions, with a mean CCA/HCC ratio between those of the two branches. These results support distinct, pathologically aligned component preferences in the dual-branch model. 
Figure~\ref{fig:tm_attention_comparison} shows a comparison of different heatmaps of a cHCC-CCA slide generated by the three-class model, branch 1, and branch 2 in the dual model. Different high attention regions correspond to the annotations.

\section{Conclusion}
\label{sec:typestyle}

We developed an attention-supervised MIL model for HCC, CCA and cHCC-CCA classification, improving the pathological plausibility of attention heatmaps while maintaining comparable classification performance. The jointly trained dual-branch model generated complementary attention maps with distinct preferences for HCC and CCA-like regions. These maps provide spatial information about tumor composition beyond a slide-level diagnosis,supporting the characterization of intratumoral heterogeneity.

Nevertheless, cHCC-CCA may exhibit distinctive morphology, and individual slides may contain only one recognizable component or regions with ambiguous differentiation. Future work will account for these characteristics and
investigate whether component-specific evidence from the dual-branch model can improve three-class classification, linking component localization with slide-level diagnosis.

\vfill
\pagebreak

\section{Compliance with ethical standards}
\label{sec:ethics}

This study was performed in line with the principles of the Declaration of Helsinki. All patients provided written informed consent. Ethical approval was granted by the Institutional Review Board “Comité d’Ethique de la Recherche (CER) Paris Nord” (IRB 00006477 of HUPNVS, Paris 7 University, AP-HP), under approval number CER-2022-169.





\bibliographystyle{IEEEbib}
\bibliography{strings,refs}

\end{document}